\documentclass[10pt,twocolumn,letterpaper]{article}

\usepackage{3dv}
\usepackage{times}
\usepackage{epsfig}
\usepackage{graphicx}
\usepackage{amsmath}
\usepackage{bm}
\usepackage{amssymb}
\usepackage{booktabs}
\usepackage{pifont}
\usepackage{multirow}
\usepackage{subcaption}
\usepackage{xcolor}
\usepackage{tabularx}
\usepackage{balance}

\newcommand\MYhyperrefoptions{bookmarks=true,bookmarksnumbered=true,
pdfpagemode={UseOutlines},plainpages=false,pdfpagelabels=true,
colorlinks=true,citecolor={black},
pdftitle={Learning Monocular Dense Depth from Events},%
pdfsubject={Deep Learning, Computer Vision, Robotics, Neuromorphic Engineering},%
pdfauthor={J. Hidalgo-Carri\'o, D. Gehrig, D. Scaramuzza},%
pdfkeywords={Deep Learning, Depth Prediction, Event Cameras}}%
\usepackage[pagebackref=true,breaklinks=true,letterpaper=true,colorlinks,bookmarks=true,\MYhyperrefoptions, pdftex]{hyperref}

\definecolor{somegray}{rgb}{0.5, 0.5, 0.5}
\newcommand{\darkgrayed}[1]{\textcolor{somegray}{#1}}
\makeatletter
\newcommand*\titleheader[1]{\gdef\@titleheader{#1}}
\AtBeginDocument{%
  \let\st@red@title\@title
  \def\@title{%
    \vskip-3em
    \bgroup\normalfont\large\centering\@titleheader\par\egroup
    \vskip1.5em\st@red@title}
}
\makeatother

\threedvfinalcopy 

\def\threedvPaperID{192} 

\ifthreedvfinal\fi

\titleheader{\darkgrayed{This paper has been accepted for publication at the\\
IEEE International Conference on 3D Vision (3DV), Fukuoka, 2020.
\copyright IEEE}}

\title{Learning Monocular Dense Depth from Events}

\begin{document}

\author{Javier Hidalgo-Carri\'o, Daniel Gehrig and Davide Scaramuzza\\
Robotics and Perception Group, University of Zurich, Switzerland\\
}

\maketitle

\begin{abstract}
    Event cameras are novel sensors that output brightness changes in the form
    of a stream of asynchronous "events" instead of intensity frames. Compared
    to conventional image sensors, they offer significant advantages: high
    temporal resolution, high dynamic range, no motion blur, and much lower
    bandwidth. Recently, learning-based approaches have been
    applied to event-based data, thus unlocking their potential and making
    significant progress in a variety of tasks, such as monocular depth
    prediction.  Most existing approaches use standard feed-forward
    architectures to generate network predictions, which do not leverage the
    temporal consistency presents in the event stream. We propose a
    recurrent architecture to solve this task and show significant improvement
    over standard feed-forward methods. In particular, our method generates
    dense depth predictions using a monocular setup, which has not been shown
    previously. We pretrain our model using a new dataset containing events
    and depth maps recorded in the CARLA simulator. We test our method on the Multi
    Vehicle Stereo Event Camera Dataset (MVSEC). Quantitative experiments show
    up to 50\% improvement in average depth error with respect to previous event-based methods.
\end{abstract}

{\small \hspace{-0.4cm}Code and dataset are available at:\newline \url{http://rpg.ifi.uzh.ch/e2depth}}


\section{Introduction}

\label{sec:intro}
Event cameras, such as the Dynamic Vision Sensor (DVS)~\cite{Lichtsteiner08ssc}
or the ATIS~\cite{Posch10isscc}, are bio-inspired vision sensors with radically
different working principles compared to conventional cameras.  While standard
cameras capture intensity images at a fixed rate, event cameras only report
changes of intensity at the pixel level and do this asynchronously at the time
they occur.  The resulting stream of events encodes the time, location, and sign
of the change in brightness.  Event cameras possess outstanding properties
when compared to standard cameras.  They have a very high dynamic range (140 dB
versus 60 dB), no motion blur, and high temporal resolution (in the order of
microseconds).  Event cameras are thus sensors that can provide high-quality
visual information even in challenging high-speed scenarios and high dynamic
range environments, enabling new application domains for vision-based
algorithms. Recently, these sensors have received great interest in various
computer vision fields, ranging from computational photography
\cite{Rebecq19pami,
Rebecq19cvpr,Scheerlinck18accv,Scheerlinck20wacv}\footnote{\url{https://youtu.be/eomALySSGVU}}
to visual odometry
\cite{Rosinol18ral,Rebecq17ral,Rebecq17bmvc,Zhu17cvpr,Zhu19cvpr,Kim14bmvc} and
depth prediction
\cite{Kim16eccv,Rebecq17ral,Rebecq16bmvc,Zhou18eccv,Zhu18eccv,Tulyakov19iccv,Zhu19cvpr}.
The survey in \cite{Gallego20pami} gives a good overview of the applications
for event cameras.

\begin{figure}[t]
    \centering
    \includegraphics[trim=0 350 0 0, clip, width=\linewidth]{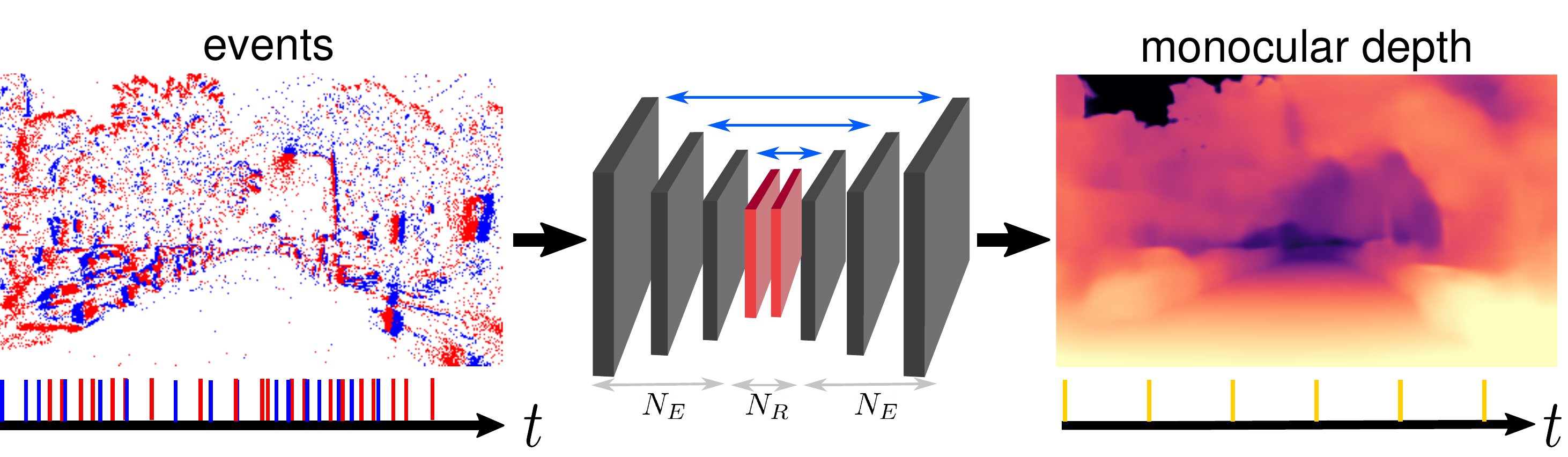}
    \caption{Method overview, the network receives asynchronous events
    inputs and predicts normalized log depth $\mathcal{\hat{D}}_k$. Our method
    uses $N_R$ recurrent blocks to leverage the temporal consistency in the
    events input.}
    \label{fig:eye_catcher}
    \vspace{-0.0cm}
\end{figure}

Monocular depth prediction has focused primarily on on
standard cameras, which work synchronously, i.e., at a fixed frame rate. State-of-the-art
approaches are usually trained and evaluated in common datasets such as
KITTI~\cite{Geiger13ijrr}, Make3D~\cite{Saxena09pami} and
NYUv2~\cite{Silberman12eccv}.

Depth prediction using event cameras has
experienced a surge in popularity in recent years~\cite{Rosinol18ral, Rebecq16bmvc,
Zhou18eccv, Rebecq17ral,Rebecq17bmvc, Kim16eccv,Zhu19cvpr, Zhou18eccv,
Zhu18eccv,Tulyakov19iccv}, due to its potential in
robotics and the automotive industry.  Event-based depth prediction is the task
of predicting the depth of the scene at each pixel in the image plane, and is
important for a wide range of applications, such as robotic grasping
\cite{Lenz15ijrr} and autonomous driving, with low-latency obstacle avoidance
and high-speed path planning.

\begin{table}[t!]
    \centering
    \resizebox{1.0\linewidth}{!} %
    {
    \begin{tabular}{c|cccc}
    \hline
        \textbf{Method}                & \textbf{Density}    &
        \textbf{Monocular} & \textbf{Metric} &
        \textbf{Learning} \\
        &  &
        & \textbf{depth} &
        \textbf{based} \\ \hline
        \cite{Rebecq17bmvc}   & \textcolor{red}{sparse}   & \textcolor{green}{yes} & \textcolor{green}{yes} & \textcolor{red}{no} \\
        \cite{Zhu17cvpr}      &\textcolor{red}{sparse}   & \textcolor{green}{yes} & \textcolor{green}{yes} & \textcolor{red}{no} \\
        \cite{Kim16eccv}      &  \textcolor{orange}{semi-dense} & \textcolor{green}{yes} & \textcolor{red}{no} &\textcolor{red}{no} \\
        \cite{Rebecq17ral}  & \textcolor{orange}{semi-dense} & \textcolor{green}{yes} & \textcolor{green}{yes} &\textcolor{red}{no} \\
        \cite{Zhou18eccv}     & \textcolor{orange}{semi-dense} & \textcolor{red}{no}  & \textcolor{green}{yes} & \textcolor{red}{no}  \\
        \cite{Zhu18eccv}      & \textcolor{orange}{semi-dense} &\textcolor{red}{no} & \textcolor{green}{yes} & \textcolor{red}{no} \\
        \hline
        \cite{Zhu19cvpr}      & \textcolor{orange}{semi-dense} & \textcolor{green}{yes} & \textcolor{green}{yes}& \textcolor{green}{yes}\\
        \cite{Tulyakov19iccv} & \textcolor{green}{dense}      & \textcolor{red}{no} & \textcolor{green}{yes} & \textcolor{green}{yes}\\
        \textbf{Ours}                  & \textcolor{green}{dense}      & \textcolor{green}{yes}       &  \textcolor{green}{yes} & \textcolor{green}{yes}    \\ \hline
    \end{tabular}
    }
    \caption{A literature review on event-based depth: model-based methods are
    listed top, learning-based methods are listed bottom. The type of output
    density is
    denoted with "sparse" (depth at pixels when only events occurred),
    "semi-dense" (depth at the reconstructed edges on the image), and "dense"
    (depth prediction at all pixels). Note than only~\cite{Tulyakov19iccv}
    addresses dense per-pixel depth, but their work uses a stereo
    setup.}
    \vspace{-0.0cm}
    \label{tab:soa}
\end{table}

However, while event-cameras have appealing properties they also present unique
challenges.  Due to the working principles of the event camera, they respond
predominantly to edges in the scene, making event-based data inherently sparse
and asynchronous.  This makes dense depth estimation with an event camera
challenging, especially in low contrast regions, which do not trigger events
and, thus, need to be filled in. Prior work in event-based depth estimation has
made significant progress in this direction, especially since the advent of deep
learning.  However, most existing works are limited: they can reliably only
predict sparse or semi-dense depth maps~\cite{Rosinol18ral, Rebecq16bmvc,
Zhou18eccv, Rebecq17ral,Rebecq17bmvc, Kim16eccv,Zhu19cvpr, Zhou18eccv,
Zhu18eccv} or rely on a stereo setup to generate dense depth predictions
\cite{Tulyakov19iccv}.

In this work, we focus on dense, monocular, and metric depth estimation using an
event camera, which addresses the aforementioned limitations. To the best of our
knowledge, this is the first time that dense monocular depth is predicted using only 
events (see Fig.~\ref{fig:eye_catcher}). We show that our approach reliably generates dense depth maps
overcoming the sparsity in a stream of events. Our methodology is based on
learning methods and gives reliable results, setting a baseline for dense depth
estimation using events. We release DENSE, a dataset recorded in CARLA, which comprises events, intensity frames, semantic labels, and depth maps. Our contributions are the following:
\begin{itemize}
   \item A recurrent network that predicts
        dense per-pixel depth from a monocular event camera.
   \item The implementation of an event camera plugin in the CARLA~\cite{Dosovitskiy17corl} simulator.
    \item DENSE - Depth Estimation oN Synthetic Events: a
       new dataset with synthetic events and perfect ground truth.
   \item Evaluation of our method on the popular Multi-Vehicle Stereo Event-Camera
       (MVSEC) Dataset \cite{Zhu18ral} where we show improved performance with
        respect to the state of the art.
\end{itemize}

\hyphenation{metric}
\hyphenation{camera}

\section{Related Work}
\label{sec:soa}

\subsection{Classical Monocular Depth Estimation }
Early work on depth prediction used probabilistic methods and feature-based
approaches. The K-means clustering approach was used by
Achanta~\emph{et~al}~\cite{Acharta12pami} to generate superpixel methods
to improve segmentation and depth. Another work proposed multi-scale features with
Markov Random Field (MRF)~\cite{Saxena06nips}. These methods tend to suffer
in uncontrolled settings, especially when the horizontal alignment  condition
does not hold.

Deep Learning significantly improved the estimate driven by convolutional
neural networks (CNN) with a variety of methods. The standard approach is to
collect RGB images with ground truth labels and train a network to predict depth
on a logarithmic scale. The network is trained in standard datasets that are
captured with a depth sensor such as laser scanning.
Eigen~\emph{et~al}~\cite{eigen2015predicting} presented the first work training
a multi-scale CNN to estimate depth in a supervised fashion. More
specifically, the architecture consists of two parts, a first estimation based on
Alexnet and a second refinement prediction. Their work led to successively major
advances in depth prediction~\cite{Godard17cvpr, godard2019digging,
wang2018learning, Li18cvpr, fu2018deep}. Better losses such as ordinal
regression, multi-scale gradient, and reverse Huber (Berhu) loss were proposed in
those works.  Another set of approaches is to jointly estimate poses and depth
in a self-supervised manner. This is the case of
Zhou~\emph{et~al}~\cite{zhou2017unsupervised}. Their work proposes to simultaneously
predict both pose and depth with an alignment loss computed from the warped
consecutive images. Most of the previous works, except for
of~\cite{Li18cvpr}, are specific for the scenario where they have been trained
and, thus, they are not domain independent.

\subsection{Event-based Depth Estimation}
Early works on event-based depth estimation used multi-view
stereo~\cite{Rebecq16bmvc} and later Simultaneous Localization and Mapping
(SLAM) \cite{Rebecq17ral,Rosinol18ral,Zhu17cvpr, Kim16eccv} to build a
representation of the environment (i.e.: map) and therefore derivate metric
depth. These approaches are model-based methods that jointly calculate pose and
map by solving a non-linear optimization problem. Model-based methods can be divided into
feature-based methods that produce sparse point clouds and direct methods that
generate semi-dense depth maps. Both methods either use
the scale given by available camera poses or rely on auxiliary sensors such as
inertial measurement units (IMU) to recover metric depth.


Purely vision-based methods have investigated the use of stereo event cameras
for depth estimation \cite{Zhou18eccv,Zhu18eccv} in which they rely on
maximizing a temporal (as opposed to photometric) consistency between the pair
of event camera streams to perform disparity and depth estimation. Recently,
several learning-based approaches have emerged that have led to significant
improvements in depth estimation \cite{Zhu19cvpr,Tulyakov19iccv}. These methods
have demonstrated more robust performance since they can integrate several cues
from the event stream. Among these, \cite{Zhu19cvpr} presents a feed-forward
neural network that jointly predicts relative camera pose and per-pixel
disparities. Training is performed using stereo event camera data, similar
to~\cite{Godard17cvpr}, and testing is done using a single input.  However, this
method still generates semi-dense depth maps, since a mask is applied to generate event frame depths at pixels where an event occurred.
The work in \cite{Tulyakov19iccv} overcomes these limitations by fusing data from
stereo setup to produce dense metric depth but still relies on a stereo setup
and a standard feed-forward architecture.  Our work compares to the
learning-based approaches but goes one step further by predicting dense metric
depth for a single monocular camera.  We achieve this by exploiting the temporal
consistency of the event stream with a recurrent convolution network
architecture and training on synthetic and real data.  Table~\ref{tab:soa}
provides a comparison of among state of the art methods, model-based and
learning-based, where our proposed approach exceeds by grouping all the listed
features.

\newcommand{\eventbatch}{\epsilon_k}
\newcommand{\eventrepr}{\textbf{E}_k}

\hyphenation{ConvLSTM}

\section{Depth Estimation Approach}
\label{sec:method}
Events cameras output events at independent pixels and do this asynchronously.
Specifically, their pixels respond to changes in the spatio-temporal log
irradiance $L(\mathbf{u},t)$ that produces a stream of asynchronous events. For
an ideal sensor, an event $e_i=(\mathbf{u}_i,t_i,p_i)$ is triggered at time
$t_i$ if the brightness change at the pixel $\mathbf{u}_i=(x_i, y_i)^\intercal$
exceeds a threshold of $\pm C$. The event polarity $p_i$ denotes the sign of
this change.

Our goal is to predict dense monocular depth from a continuous stream of events.
The method works by processing subsequent non-overlapping windows of events
$\eventbatch=\{e_i\}_{i=0}^{N-1}$ each spanning a fixed interval $\Delta T =
t^k_{N-1}-t^k_0$.  For each window, we predict log depth maps
$\{\mathcal{\hat{D}}_k\}$, with $\mathcal{\hat{D}}_k\in [0,1]^{W\times H}$.  We
implement log depth prediction as a recurrent convolutional neural network with
an internal state $\mathbf{s}_k$. We train our network in a supervised manner,
using ground truth depth maps. The network is first trained in simulation using
perfect ground truth and synthetic events and finetuned in a real
sequence.

\subsection{Event Representation}
Due to the sparse and asynchronous nature of event data, batches of events
$\eventbatch$ need to be converted to tensor-like representations $\eventrepr$.
One way to encode these events is by representing them as a spatio-temporal
voxel grid \cite{Zhu19cvpr,Gehrig19iccv} with dimensions $B \times H \times W$.
Events within the time window $\Delta T$ are collected into $B$ temporal bins
according to

\begin{equation}
    \eventrepr(\mathbf{u}_k,t_n)=\sum_{e_i}p_i \delta(\mathbf{u}_i-\mathbf{u}_k)\max(0,1-\vert t_n - t^*_i\vert)
\end{equation}

where $t_i^*=\frac{B-1}{\Delta T}(t_i-t_0)$ is the normalized event timestamp.
In our experiments, we used $\Delta T = 50 ms$ of events and $B=5$ temporal bins.
To facilitate learning, we further normalize the non-zero values in the voxel
grid to have zero mean and unit variance.

\subsection{Network Architecture}

\begin{figure}[t!]
    \centering
    \includegraphics[width=\linewidth]{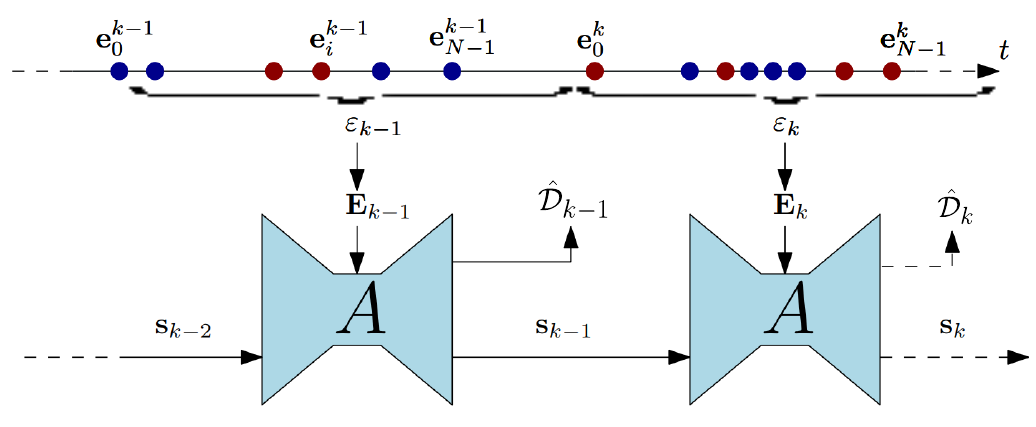}\\
    \caption{Our network architecture, image adapted
    from~\cite{Rebecq19pami}. The event stream is grouped into non-overlapping
    windows of events and converted to tensor-like voxel grids~\cite{Zhu19cvpr}.
    These voxel grids are passed to our recurrent fully convolutional neural
    network to produce normalized log depth predictions.}
    \label{fig:arch}
\end{figure}

It consists of a recurrent, fully convolutional neural network, based on the UNet
architecture \cite{Ronneberger15icmicci}.  The network input is first processed
by a head layer $\mathcal{H}$ and then $N_E$ recurrent encoder layers
($\mathcal{E}^i)$ followed by $N_R$ residual blocks ($\mathcal{R}^j$) and $N_E$
decoder layers $\mathcal{D}^l$.  A final depth prediction layer $\mathcal{P}$
produces the output of our network.   The head layer produces an output with
$N_b$ channels, which is doubled at each encoder layer, resulting in a feature
map with $N_b \times 2^{N_E}$ output channels.~$\mathcal{P}$ performs a
depth-wise convolution with one output channel and kernel size 1. We use skip
connections between symmetric encoder and decoder layers (see
Fig.~\ref{fig:arch}).  At the final layer, the activations are squashed by a
sigmoid activation function.  Each encoder layer is composed of a downsampling
convolution with kernel size 5 and stride 2 and a ConvLSTM \cite{Shi15nips}
module with kernel size 3. The encoding layers maintain a state $c_k^i$ which is
at 0 for $k=0$.  The residual blocks use a kernel size of 3 and apply summation
over the skip connection.  Finally, each decoder layer is composed of a bilinear
upsampling operation followed by convolution with kernel size 5.  We use ReLU
except for the prediction layer, and batch normalization \cite{Ioffe15icml}.  In
this work we use $N_E=3$, $N_R=2$ and $N_b=32$ and we unroll the network for
$L=40$ steps.

\subsection{Depth Map Post-processing}
As usual, in recent work on depth prediction, we train our network to predict
a normalized log depth map.  Log depth maps have the advantage of representing
large depth variations in a compact range, facilitating learning.  If
$\mathcal{\hat{D}}_k$ is the depth predicted by our model, the metric depth can
be recovered by performing the following operations:

\begin{equation}
    \mathcal{\hat{D}}_{m,k} = \mathcal{D}_{\text{max}}\exp(-\alpha
    (1-\mathcal{\hat{D}}_{k}))
\end{equation}

Where $\mathcal{D}_{\text{max}}$ is the maximum expected depth and $\alpha$ is a
parameter chosen, such that a depth value of 0 maps to minimum observed depth.
In our case, $\mathcal{D}_{\text{max}}=80$ meters and $\alpha=3.7$ corresponding
to a minimum depth of $~2$ meters.

\subsection{Training Details}
We train our network in a supervised fashion, by minimizing the scale-invariant
and multi-scale scale-invariant gradient matching losses at each time step.
Given a sequence of ground truth depth maps $\{\mathcal{D}_k\}$, denote the
residual $\mathcal{R}_k=\mathcal{\hat{D}}_{k}-\mathcal{D}_{k}$.  Then the
scale-invariant loss is defined as

\begin{equation}
    \mathcal{L}_{k,\text{si}} =
    \frac{1}{n}\sum_\textbf{u}(\mathcal{R}_k(\textbf{u}))^2-\frac{1}{n^2}\left(\sum_\textbf{u} \mathcal{R}_k(\textbf{u})\right)^2,
\end{equation}

where $n$ is the number of valid ground truth pixels $\textbf{u}$.  The
multi-scale scale-invariant gradient matching loss encourages smooth depth
changes and enforces sharp depth discontinuities in the depth map prediction. It
is computed as follows:

\begin{equation}
    \mathcal{L}_{k,\text{grad}}=\frac{1}{n}\sum_s \sum_\textbf{u} \vert \nabla_x \mathcal{R}^s_k(\textbf{u})\vert + \vert \nabla_y \mathcal{R}^s_k(\textbf{u}) \vert.
\end{equation}

Here $\mathcal{R}^s_k(\textbf{x})$ refers to the residual at scale $s$ and the
$L_1$ norm is used to enforce sharp depth discontinuities in the prediction.  In
this work, we consider four scales, similar to \cite{Li18cvpr}.  The resulting
loss for a sequence of $L$ depth maps is thus

\begin{equation}
    \mathcal{L}_{\text{tot}} = \sum_{k=0}^{L-1} \mathcal{L}_{k,\text{si}}+\lambda \mathcal{L}_{k,\text{grad}}.
\end{equation}

The hyper-parameter $\lambda = 0.5$ was chosen by cross-validation. We train
with a batch size of 20 and a learning rate of $10^{-4}$ and use the Adam
\cite{Kingma15iclr} optimizer.

Our network requires training data in the form of events sequences with
corresponding depth maps. However, it is difficult to get perfect dense ground
truth depth maps in real datasets. For this reason, we propose to first train the
network using synthetic data and get the final metric scale by finetuning the
network using real events from the MVSEC dataset.

We implement an event camera sensor in CARLA~\cite{Dosovitskiy17corl} based on
the previous event simulator ESIM~\cite{Rebecq18corl}. The event camera sensor
takes the rendered images from the simulator environment and computes
per-pixel brightness change to simulate an event camera. The computation is done
at a configurable but fixed high framerate (we use $20$ times higher than the images
frame rate) to approximate the continuous signal of a real event
camera. The simulator allows us to capture a variety of scenes with
different weather conditions and illumination properties. The camera parameters
are set to mimic the event camera at MVSEC with a sensor size of $346 \times
260$ pixels (resolution of the DAVIS346B) and a focal length of $83^\circ$
horizontal field of view.

We split DENSE, our new dataset with synthetic events, into five sequences for
training, two sequences for validation, and one sequence for testing (a total of
eight sequences). Each sequence consists of 1000 samples at $30$ FPS
(corresponding to $33$ seconds), each sample is a tuple of one RGB image, the
stream of events between two consecutive images, ground truth depth, and
segmentation labels. Note that only the events and the depth, maps are used by
the network for training. RGB images are provided for visualization and
segmentation labels complete the dataset with richer information. CARLA Towns 01
to 05 are the scenes for training, Town 06 and 07 for validation, and the test
sequence is acquired using Town 10. This split results in an overall of $5000$
samples for training, $2000$ samples for validation, and $1000$ samples for
testing.

\section{Experiments}
\label{sec:results}

In this section, we present qualitative and quantitative results and compare them with
previous methods~\cite{Zhu19cvpr} on the MVSEC dataset. We focus our evaluation
on real event data while the evaluation on synthetic data is detailed in
Appendix A.

\begin{table*}[t!]
    \centering
    \resizebox{1.0\textwidth}{!} %
    {
        \begin{tabular}{l | c | c  c c c c|c c c }
        \toprule
             \textbf{Training set} & \textbf{Dataset} &
             \textbf{Abs Rel}$\downarrow$ &  \textbf{Sq Rel} $\downarrow$ &
             \textbf{RMSE}$\downarrow$ &  \textbf{RMSE log}$\downarrow$ &
             \textbf{SI log}$\downarrow$ & $ \boldsymbol{\delta < 1.25\uparrow}$ &
             $\boldsymbol{\delta < 1.25^2 \uparrow}$  & $\boldsymbol{\delta < 1.25^3 \uparrow}$ \\ 
        \toprule
        S &                          & 0.698& 3.602 & 12.677 &  0.568& 0.277 & 0.493 & 0.708 & 0.808\\
        R &  outdoor day1 & 0.450& 0.627 & 9.321 &  0.514& 0.251 & 0.472 & 0.711 & 0.823\\
        S* $\rightarrow$ R &  & 0.381& \textbf{0.464} & 9.621 &  0.473& 0.190 & 0.392 & 0.719 & 0.844\\
        S* $\rightarrow$ (S+R) &  & \textbf{0.346} & 0.516 & \textbf{8.564} & \textbf{0.421}& \textbf{0.172} & \textbf{0.567} & \textbf{0.772} & \textbf{0.876}\\
        \hline
        S &                              & 1.933 & 24.64 & 19.93 &  0.912& 0.429 & 0.293  & 0.472 & 0.600\\
        R & outdoor night1  & 0.770 & 3.133 & \textbf{10.548} &  0.638& 0.346 & 0.327  & 0.582 & 0.732\\
        S* $\rightarrow$ R &  & \textbf{0.554} & \textbf{1.798} & 10.738 &  \textbf{0.622}& \textbf{0.343} & 0.390  & 0.598 & 0.737\\
        S* $\rightarrow$ (S+R) &  & 0.591 & 2.121 & 11.210 &  0.646& 0.374 &
            \textbf{0.408}  & \textbf{0.615} & \textbf{0.754}\\
        \hline
        S &                            & 0.739 & 3.190 & 13.361 &  0.630& 0.301 & 0.361  & 0.587 & 0.737\\
            R & outdoor night2 & 0.400 & 0.554 & \textbf{8.106} &  \textbf{0.448}& \textbf{0.176} & 0.411  & 0.720 & 0.866\\
        S* $\rightarrow$ R &  & 0.367 & \textbf{0.369} & 9.870 &  0.621& 0.279 & 0.422  & 0.627 & 0.745\\
        S* $\rightarrow$ (S+R) &  & \textbf{0.325} & 0.452 & 9.155  &
            0.515& 0.240 & \textbf{0.510}  & \textbf{0.723} & \textbf{0.840}\\
        \hline
        S &                               & 0.683 & 1.956 & 13.536 &  0.623& 0.299 & 0.381  & 0.593 & 0.736\\
            R &   outdoor night3 & 0.343 & 0.291 & \textbf{7.668} &
            \textbf{0.410} & \textbf{0.157} & 0.451  & 0.753 & \textbf{0.890}\\
        S* $\rightarrow$ R &  & 0.339 & 0.230 & 9.537 &  0.606& 0.258 & 0.429  & 0.644 & 0.760\\
        S* $\rightarrow$ (S+R) &  & \textbf{0.277} & \textbf{0.226} & 8.056 &  0.424 & 0.162 & \textbf{0.541}  & \textbf{0.761} & \textbf{0.890}\\
        \bottomrule
        \end{tabular}%
    }
    \caption{Ablation study and evaluation of MVSEC. All rows are the same
    network with the change in the training set.  The Training set is denoted with $S$
    (synthetic data from the DENSE training split), $R$ (real data from the
    training split in~\textit{outdoor day2} sequence), $S^*$ (first 1000 samples of the
    DENSE training split), $S^*\rightarrow R$ (pretrained on $S^*$ and retrained on
    $R$), $S^*\rightarrow (S+R)$ (pretrained on $S^*$ and retrained on both
    datasets). $\downarrow$ indicates lower is better and $\uparrow$ higher is
    better. The results are the driving sequences of MVSEC (except
    for~\textit{outdoor day2}). Best values are shown in bold.}
    \label{tab:ablation_metrics}
\end{table*}

\begin{figure*}[t!]
    \centering
    $\begin{array}{ccc}
    \hspace{-0.12cm}
    \subcaptionbox{\label{fig:ablation_events} Events}
    {
        \includegraphics[width=0.33\textwidth,keepaspectratio]{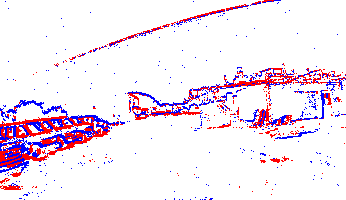}
    }
    \hspace{-0.1cm}
    \subcaptionbox{\label{fig:simu} Depth trained only on $S$ data}
    {
        \includegraphics[width=0.33\textwidth,keepaspectratio]{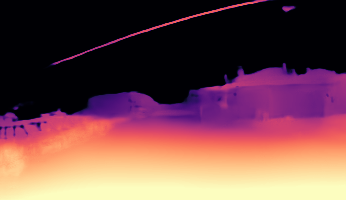}
    }
    \hspace{-0.1cm}
        \subcaptionbox{\label{fig:real} Depth trained only on $R$ data}
    {
        \includegraphics[width=0.33\textwidth,keepaspectratio]{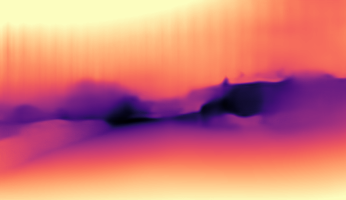}
    }
    \end{array}$

    $\begin{array}{ccc}
    \hspace{-0.12cm}
    \subcaptionbox{\label{fig:simu_real} Depth trained on $S^*\rightarrow R$ }
    {
        \includegraphics[width=0.33\textwidth,keepaspectratio]{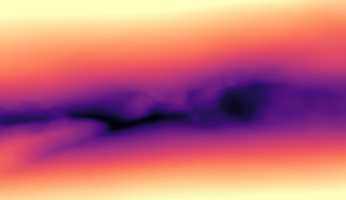}
    }
    \hspace{-0.1cm}
    \subcaptionbox{\label{fig:mixed} Depth trained on $S^*\rightarrow (S+R)$ }
    {
        \includegraphics[width=0.33\textwidth,keepaspectratio]{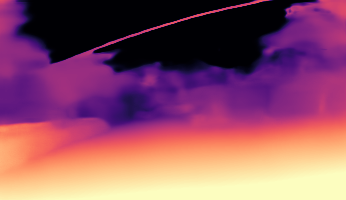}
    }
    \hspace{-0.1cm}
        \subcaptionbox{\label{fig:ablation_gt} Depth ground truth }
    {
        \includegraphics[width=0.33\textwidth,keepaspectratio]{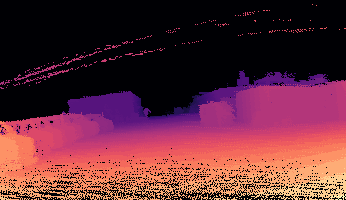}
    }
   \end{array}$
    \caption{Ablation study of our method trained with different training sets
    (see Table~\ref{tab:ablation_metrics}). Fig.~\ref{fig:ablation_events} shows
    the events, from Fig.~\ref{fig:simu} to Fig.~\ref{fig:mixed} the predicted
    dense monocular depth using different training sets.
    Fig.~\ref{fig:ablation_gt} depicts the corresponding ground truth. The depth
    maps are shown in logarithmic scale and correspond to sample $3562$ in
    the~\textit{outdoor day1} sequence of MVSEC.}
   \label{fig:ablation_visual}
\end{figure*}

\begin{table*}[t!]
\centering
\resizebox{1.0\textwidth}{!} %
{
\begin{tabular}{c|c|c c c|c c c c c}
\toprule
    \multirow{2}{*}{\textbf{Dataset}}     & \multirow{2}{*}{\textbf{Distance}} &
    \multicolumn{3}{c|}{\textbf{Frame based}}   & \multicolumn{5}{c}{\textbf{Event based}} \\ 
                             &                          & MonoDepth~\cite{Godard17cvpr}  & MegaDepth~\cite{Li18cvpr}
                             & MegaDepth$^{+}$~\cite{Li18cvpr} & Zhu et
                             al.~\cite{Zhu19cvpr}  & Ours$^{S}$ & Ours$^{R}$&
                             Ours$^{S^* \rightarrow R}$ & Ours$^{\#}$\\

    \toprule
                                 & 10\text{m}                  &   3.44        &   2.37        &    3.37        &     2.72      &     4.60        &    2.70         &     2.13         &   \textbf{1.85}   \\ 
    \multirow{1}{*}{outdoor day1}                                 & 20\text{m}          &   7.02        &   4.06        &    5.65        &     3.84      &      5.66       &     3.46        &     2.68        &  \textbf{2.64}    \\
                                 & 30\text{m}        &     10.03      &    5.38       &     7.29       &    4.40       &      6.10       &     3.84        &     3.22         &  \textbf{3.13}   \\
    \hline
                              & 10\text{m}                   &    3.49       &     2.54      &    \textbf{2.40}        &   3.13     &      10.36       &    5.36      &    3.31        &   3.38   \\ 
    \multirow{1}{*}{outdoor night1}                                 & 20\text{m}         &    6.33       &     4.15      &    4.20        &    4.02       &      12.97       &        5.32     &     \textbf{3.73}         &  3.82    \\
                                 & 30\text{m}         &    9.31       &    5.60       &      5.80      &       4.89    &      13.64       &     5.40        &     \textbf{4.32}         &   4.46   \\
    \hline
                          & 10\text{m}                  &      5.15     &      3.92     &       3.39     &       2.19    &      6.14       &     2.80        &      1.99        &    \textbf{1.67}  \\ 
    \multirow{1}{*}{outdoor night2}                                 & 20\text{m}        &   7.80        &    5.78       &    4.99        &      3.15     &     8.64        &    3.28         &    3.14          &   \textbf{2.63}   \\
                                 & 30\text{m}         &    10.03       &    7.05       &     6.22       &    3.92      &     9.57        &     3.74        &     4.14         &  \textbf{3.58}    \\
    \hline
                      & 10\text{m}                   &    4.67       &     4.15      &     4.56       &     2.86      &      5.72       &     2.39        &    1.76          &   \textbf{1.42}   \\ 
    \multirow{1}{*}{outdoor night3}                                 & 20\text{m}         &   8.96        &      6.00     &     5.63       &    4.46       &     8.29        &     2.88        &    2.98          &  \textbf{2.33}    \\
                                 & 30\text{m}          &   13.36        &    7.24       &     6.51      &     5.05      &     9.27        &    3.39         &    3.98          &    \textbf{3.18}  \\
    \bottomrule
\end{tabular}
}
\caption{Average absolute depth errors (in meters) at different
    cut-off depth distances (lower is better). MegaDepth$^{+}$ refers to
    MegaDepth~\cite{Li18cvpr} using E2VID~\cite{Rebecq19pami} reconstructed
    frames and Ours$^{\#}$ refers to our method trained using
    $S^* \rightarrow (S+R)$. Our results outperform state of the art image-based
monocular depth prediction methods~\cite{Godard17cvpr, Li18cvpr} while
outperforming state of the art event-based methods~\cite{Zhu19cvpr}.}
\label{tab:results}
\end{table*}

\begin{figure*}[t!]
    \centering
    $\begin{array}{cccc}
    {
        \includegraphics[width=0.25\textwidth,keepaspectratio]{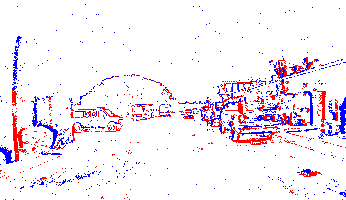}
    }
    \hspace{0.05cm}
    {
        \includegraphics[width=0.25\textwidth,keepaspectratio]{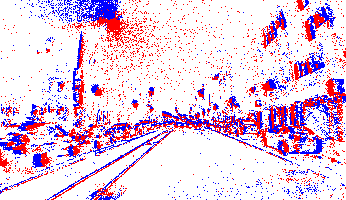}
    }
    \hspace{0.05cm}
    {
        \includegraphics[width=0.25\textwidth,keepaspectratio]{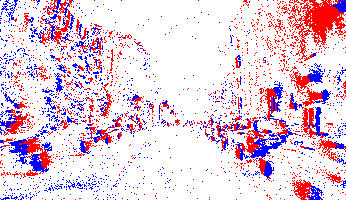}
    }
    \hspace{0.05cm}
    {
        \includegraphics[width=0.25\textwidth,keepaspectratio]{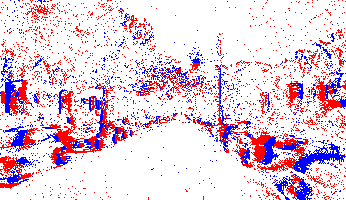}
    }
    \end{array}$

    $\begin{array}{cccc}
    {
        \includegraphics[width=0.25\textwidth,keepaspectratio]{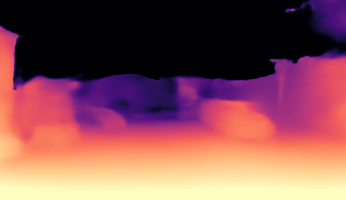}
    }
    \hspace{0.05cm}
    {
        \includegraphics[width=0.25\textwidth,keepaspectratio]{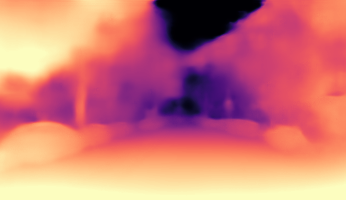}
    }
    \hspace{0.05cm}
    {
        \includegraphics[width=0.25\textwidth,keepaspectratio]{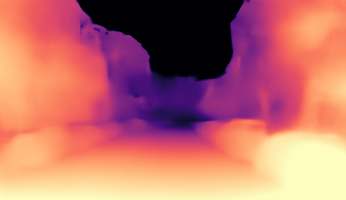}
    }
    \hspace{0.05cm}
    {
        \includegraphics[width=0.25\textwidth,keepaspectratio]{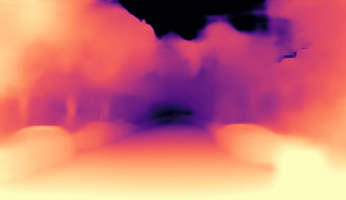}
    }
    \end{array}$

    $\begin{array}{cccc}
    \hspace{0.05cm}
    {
        \includegraphics[width=0.25\textwidth,keepaspectratio]{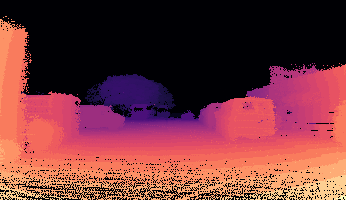}
    }
    \hspace{0.05cm}
    {
        \includegraphics[width=0.25\textwidth,keepaspectratio]{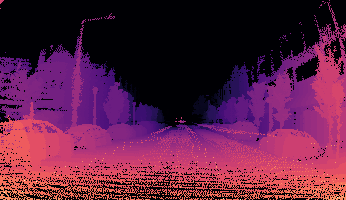}
    }
    \hspace{0.05cm}
    {
        \includegraphics[width=0.25\textwidth,keepaspectratio]{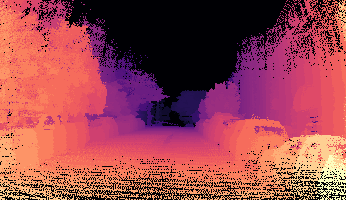}
    }
    \hspace{0.05cm}
    {
        \includegraphics[width=0.25\textwidth,keepaspectratio]{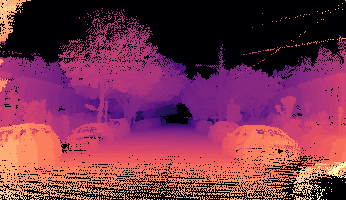}
    }
    \end{array}$

    $\begin{array}{cccc}
    {
        \includegraphics[width=0.25\textwidth,keepaspectratio]{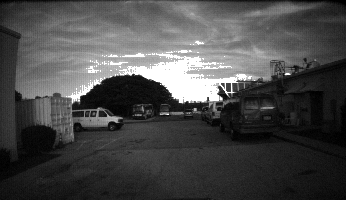}
    }
    \hspace{0.05cm}
    {
        \includegraphics[width=0.25\textwidth,keepaspectratio]{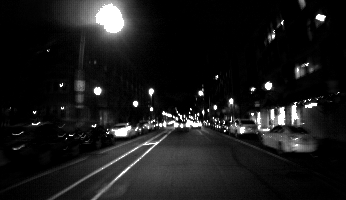}
    }
    \hspace{0.05cm}
    {
        \includegraphics[width=0.25\textwidth,keepaspectratio]{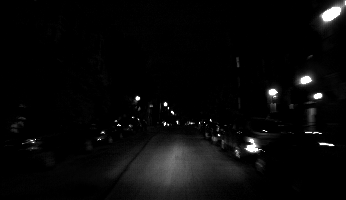}
    }
    \hspace{0.05cm}
    {
        \includegraphics[width=0.25\textwidth,keepaspectratio]{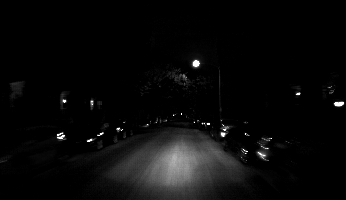}
    }
    \end{array}$

    $\begin{array}{cccc}
    \hspace{-0.12cm}
    \subcaptionbox{\label{fig:outdoor_day1} outdoor day1}
    {
        \includegraphics[width=0.25\textwidth,keepaspectratio]{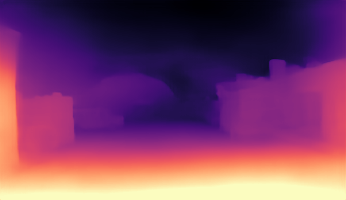}
    }
    \hspace{-0.12cm}
    \subcaptionbox{\label{fig:outdoor_night1} outdoor night1}
    {
        \includegraphics[width=0.25\textwidth,keepaspectratio]{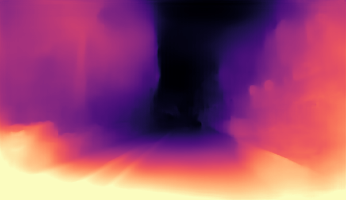}
    }
    \hspace{-0.12cm}
    \subcaptionbox{\label{fig:outdoor_night2} outdoor night2}
    {
        \includegraphics[width=0.25\textwidth,keepaspectratio]{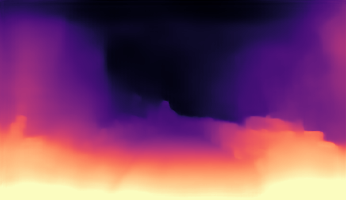}
    }
    \hspace{-0.12cm}
    \subcaptionbox{\label{fig:outdoor_night3} outdoor night3}
    {
        \includegraphics[width=0.25\textwidth,keepaspectratio]{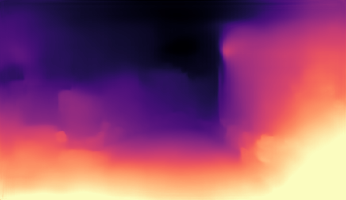}
    }
    \end{array}$

    \caption{Qualitative comparison of the four test sequences of MVSEC dataset.
    The first row shows the events. Second row our dense depth
    map predictions. Third row the ground truth depth maps. The Fourth row shows
    DAVIS grayscale frames and the fifth row the MegaDepth prediction using the
    grayscale frames.~\textit{outdoor day1} corresponds to sample $3741$ in the
    sequence.~\textit{outdoor night1} to sample $288$ in the sequence.
    ~\textit{outdoor night2} corresponds to sample $2837$ in the sequence
    and~\textit{outdoor night3} to sample $2901$ in the sequence.}
   \label{fig:results_visual}
\end{figure*}

We perform an ablation study by training on synthetic and real events
to show the benefits of using synthetic training data from the simulation. We train
our recurrent network using the training split of the DENSE dataset (5000
samples) and train for $300$ epochs (75000 iterations). We convert the events to
a voxel grid synchronized with the ground truth depth which is 30
FPS in simulation. We use each depth image to supervise the training by first converting the
absolute depth to a logarithmic scale. A sample is formed by a voxel grid of
events between two consecutive frames and the ground truth label. Consecutively,
the network is also trained with real data using the
MVSEC~\footnote{\url{https://daniilidis-group.github.io/mvsec/}} dataset in the same
manner. We unpack the available online data in ROS bag format and pack the
stream of events in voxel grid synchronized with the ground truth depth which
is 20 FPS.  We use~\textit{outdoor day2} sequence of MVSEC to train the network
as in~\cite{Zhu19cvpr}. More specifically we split the sequence into training,
validation, and testing. The training split consist of 8523 samples, the
validation split contains 1826 samples and the test split comprise the rest with
another 1826 samples. We also train for 300 epochs in the real data (127800
iterations). We perform data augmentation, in synthetic and real data, by random
cropping and horizontal flip of the training samples. The network and data loader
are implemented in Pytorch.

The quantitative results are shown in Table~\ref{tab:ablation_metrics}, and are
supported with qualitative depth maps in Fig~\ref{fig:ablation_visual}. The
network predicts depth in the logarithmic scale, which is normalized and
restored to absolute values by multiplying by the maximum depth clipped at
$80$~\text{m}. We compare the results for different combinations of training
sets. We first show the results by training only with synthetic data $S$.
Afterward,  we show that training in real data $R$ drastically improve the
metrics with respect to only synthetic data $S$. The incorporation of real
events improves all metrics, especially the Absolute relative error - Abs. Rel,
which is the most informative value. This is because the network is capable of
closing the domain gap when seeing data from the real world. We also notice that
pretraining the network with a subset ($1000$ samples) of the synthetic
data $S^*$ and then training in real data helps the network to converge faster
(first $100$ epochs).  This is depicted by $S^* \rightarrow R$ in
Table~\ref{tab:ablation_metrics} and the results increase the performance in
almost all values with respect to training only with real data $R$. The penalty
of training with real data is in the qualitative depth map (see
Fig.~\ref{fig:real} and~\ref{fig:simu_real}). This is because having perfect
align ground truth data with events is hard to obtain in the real world.  The
lack of perfect ground truth prevents the network to predict depth with sharp
edges and have difficulties to mask the sky. For that reason, we mixed synthetic
and real data, denoted by $(S+R)$, to fine tune the network and get the
best of both. Training with real reduces the errors by predicting
the correct metric while using synthetic data helps to estimate qualitatively
better depth maps. The combination of synthetic and real training data is
possible without changing the loss function since both datasets mimic the
DAVIS346B sensor resolution and focal length.

The ablation study shows that synthetic data enhance the results. It also shows
that the potential of monocular depth prediction grows with respect to the
amount of training data. We now further compare our method against the state of the
art monocular depth: two image-based techniques, MonoDepth~\footnote{MonoDepth
performs more accurately than MonoDepth2 for the MVSEC dataset}~\cite{Godard17cvpr}
and MegaDepth~\cite{Li18cvpr}, and the event-based approach
from~\cite{Zhu19cvpr} (see Table~\ref{tab:results}). MegaDepth is further
applied to frames reconstructed from events using E2VID~\cite{Rebecq19pami}. The
evaluation is done using the average mean error at depths of $10$\text{m},
$20$\text{m}, and $30$\text{m} since these are the available metrics reported
until now in the MVSEC dataset. The values for MonoDepth are directly taken from
the evaluation in~\cite{Zhu19cvpr}. Our work gives more accurate depth
prediction at all distances with an average improvement overall sequence of
26.25\% at 10\text{m}, 25.25\% at 20\text{m} and 21.0\% at 30\text{m} with
respect to values reported in~\cite{Zhu19cvpr}.  Our method produces dense depth
results up to 50.0\% improvement with respect to previous methods in
\textit{outdoor night3} sequence of MVSEC. Image-based methods have difficulties
to predict depth in low light conditions. MegaDepth applied to reconstructed
frames performs more accurately in night sequences. However, the direct use of
events (i.e.: end to end without parsing through image reconstruction) in our
method gives a better estimate since the events capture increments in contrast
at a higher temporal resolution.  Appendix B further explains this fact in a
high dynamic range situation. In all the night sequences our method outperforms
previous approaches except for \textit{outdoor night1} at
$10$\text{m}. This is because~\textit{outdoor night1}, compared to other
sequences, has a higher amount of moving objects in front of the car. This
creates spurious measurements in the path of such objects which have not been
removed from the ground truth depth in the dataset.

None of the methods uses samples from night driving sequences at training time,
neither image-based methods nor event-based solutions. MegaDepth is trained with
the MD dataset from images available on the Internet and this achieves superior
generalizability than MonoDepth which is trained with KITTI~\cite{Geiger13ijrr}
and reported the values from Zhu et al~\cite{Zhu19cvpr}.
Fig.~\ref{fig:results_visual} contains a visual comparison. Each column
corresponds to an MVSEC sequence
(Fig.~\ref{fig:outdoor_day1}-\ref{fig:outdoor_night3}). The first and second row
depict DAVIS frames and depth prediction from MegaDepth using standard frames.
The third row shows the events warped to a frame at the same synchronized
sample. The last two rows are our dense prediction and ground truth. It can be
noticed that~\textit{outdoor night2} and~\textit{outdoor night3} are
particularly dark scenes making predictions a challenging task for conventional
image-based methods. Our method better preserves the car shapes at both sides of
the road, while they are completely neglected by standard images and therefore
omitted in the prediction from MegaDepth.

\begin{table*}[t!]
    \centering
    \resizebox{1.0\textwidth}{!} %
    {
        \begin{tabular}{l | c  c c c c|c c c | c c c }
        \toprule
            \textbf{Dataset} &
             \textbf{Abs Rel}$\downarrow$ &  \textbf{Sq Rel} $\downarrow$ &
             \textbf{RMSE}$\downarrow$ &  \textbf{RMSE log}$\downarrow$ &
             \textbf{SI log}$\downarrow$ & $ \boldsymbol{\delta < 1.25\uparrow}$ &
             $\boldsymbol{\delta < 1.25^2 \uparrow}$  & $\boldsymbol{\delta <
             1.25^3 \uparrow}$ & \textbf{Avg. error 10\text{m}$ \downarrow$} &
             \textbf{Avg. error 20\text{m}$\downarrow$}  & \textbf{Avg. error 30\text{m}$\downarrow$ } \\
        \toprule
        Town06 & 0.120 & 0.083 & 6.640 &  0.188 & 0.035 & 0.855  & 0.956 &
            0.987 & 0.31 & 0.74 & 1.32\\
        Town07 & 0.267 & 0.535 & 10.182 &  0.328 & 0.098 & 0.774  & 0.878 &
            0.927 & 1.03 & 2.35 & 3.06\\
        \hline
        Town10 & 0.220 & 0.279 & 11.812 &  0.323 & 0.093 & 0.724 & 0.865 &
            0.932 & 0.61 & 1.45 & 2.42\\

        \bottomrule
        \end{tabular}%
    }
    \caption{Quantitative results on the DENSE dataset. We train the network only on
    synthetic events from the training split $S$. The first two sequences are
    used for validation and the Town10 sequence for testing.}
    \label{tab:carla_metrics}

\end{table*}

\begin{figure*}[t!]
    \centering
    $\begin{array}{cccc}
    {
        \includegraphics[width=0.25\textwidth,keepaspectratio]{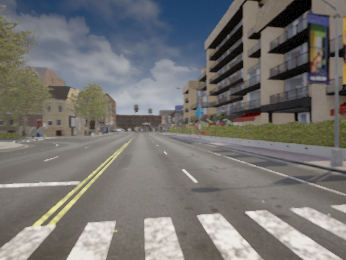}
    }
    \hspace{0.05cm}
    {
        \includegraphics[width=0.25\textwidth,keepaspectratio]{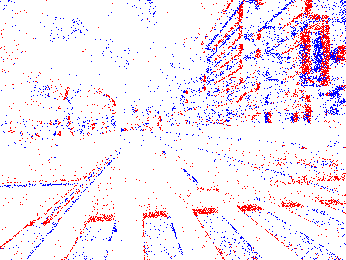}
    }
    \hspace{0.05cm}
    {
        \includegraphics[width=0.25\textwidth,keepaspectratio]{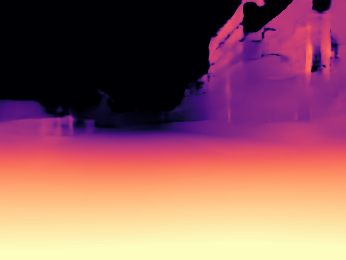}
    }
    \hspace{0.05cm}
    {
        \includegraphics[width=0.25\textwidth,keepaspectratio]{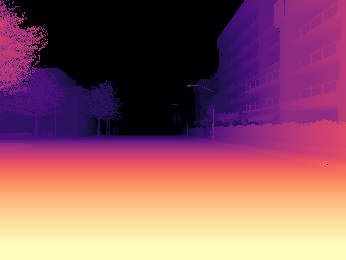}
    }
    \end{array}$

    $\begin{array}{cccc}
    \hspace{-0.12cm}
    \subcaptionbox{\label{fig:icarla_rgb} Frame}
    {
        \includegraphics[width=0.25\textwidth,keepaspectratio]{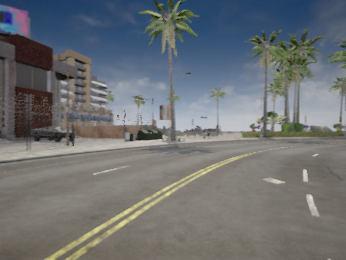}
    }
    \hspace{-0.12cm}
    \subcaptionbox{\label{fig:carla_events} Events}
    {
        \includegraphics[width=0.25\textwidth,keepaspectratio]{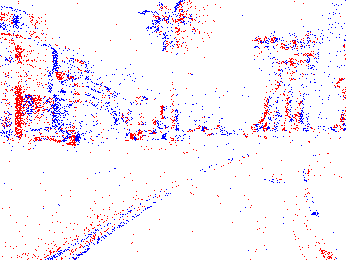}
    }
    \hspace{-0.12cm}
    \subcaptionbox{\label{fig:carla _prediction} Ours on events}
    {
        \includegraphics[width=0.25\textwidth,keepaspectratio]{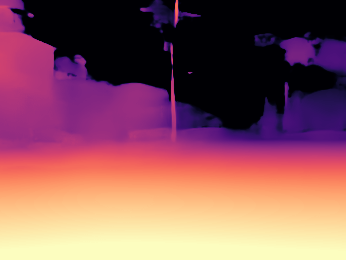}
    }
    \hspace{-0.12cm}
    \subcaptionbox{\label{fig:carla_gt} Ground truth}
    {
        \includegraphics[width=0.25\textwidth,keepaspectratio]{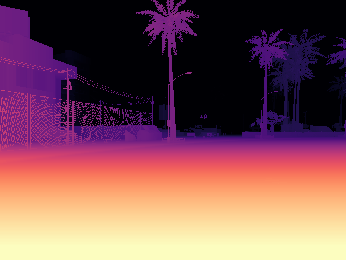}
    }

    \end{array}$

    \caption{Qualitative results on DENSE for the Town10 sequence. The first row
    corresponds to sample $143$ and the second row to sample $547$ in the
    sequence.}
   \label{fig:carla_results}
    \vspace{-0.5cm}
\end{figure*}

\hyphenation{monocular}

\section{Conclusion}
\label{sec:conclusion}

In this paper, we presented the first work on monocular dense estimation from events. Our
solution exploits the benefits of recurrent convolutional neural networks to infer dense depth from a stream of asynchronous events. We reported
results on the Multi Vehicle Stereo Event Camera Dataset (MVSEC) which is the
only currently public dataset comprised of events, frames, and ground-truth depth. We showed that training on synthetic data is beneficial
for several reasons: it helps the network converge faster, depth maps have
a better quality due to perfect ground truth, and simulation captures a larger
variety of conditions. Finally, we showed that our methodology produces dense
depth maps with more accuracy than existing methods.

\section{Acknowledgments}
\label{sec:ack}

This work was supported by Prophesee, the Swiss National Center of Competence in
Research Robotics (NCCR), through the Swiss National Science Foundation, and the
SNSF-ERC starting grant. The authors would like to thank Alessio Tonioni
(Google Zurich) and Federico Tombari (TU Munich) for their valuable
feedback and insights to accomplish this work.

\appendix
\section{Results on Synthetic Data}
\label{sec:appendixa}

We show quantitative and qualitative results of our method in the DENSE
dataset. Fig~\ref{fig:carla_results} shows the qualitative results on the DENSE
test sequence corresponding with Town10 in CARLA. We also show the quantitative
results for two validation datasets, Town06, and Town07 (see
Table~\ref{tab:carla_metrics}).  The results show metric numbers within the
range of state of the art image-based methods in popular datasets like KITTI.
This emphasizes our statement that events have enough information to estimate
dense monocular depth.

\section{Why using events for Depth prediction?}
\label{sec:appendixb}

\begin{figure*}[t!]
    \centering
    $\begin{array}{cccc}
    \subcaptionbox{\label{fig:tunnel_frame} Frame}
    {
        \includegraphics[width=0.25\textwidth,keepaspectratio]{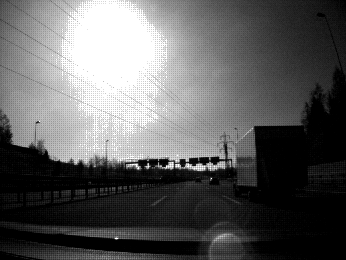}
    }
    \hspace{-0.1cm}
    \subcaptionbox{\label{fig:tunnel_megadepth} MegaDepth on frames~\cite{Li18cvpr}}
    {
        \includegraphics[width=0.25\textwidth,keepaspectratio]{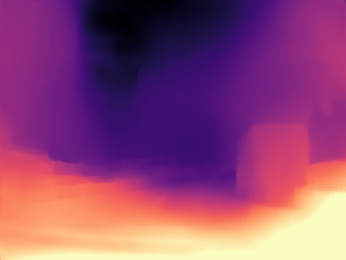}
    }
    \hspace{-0.1cm}
    \subcaptionbox{\label{fig:tunnel_events} Events}
    {
        \includegraphics[width=0.25\textwidth,keepaspectratio]{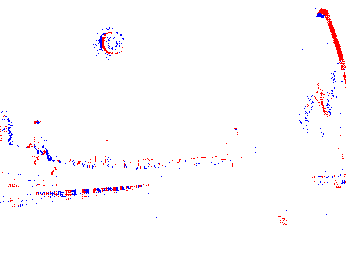}
    }
    \hspace{-0.1cm}
    \subcaptionbox{\label{fig:tunnel_prediction} Ours on events}
    {
        \includegraphics[width=0.25\textwidth,keepaspectratio]{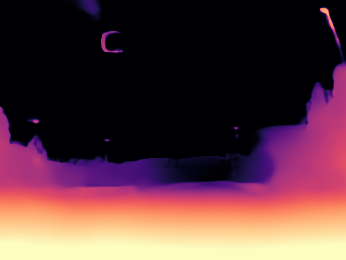}
    }
    \end{array}$
    \caption{Qualitative results in a high dynamic range (HDR) situation facing
    the sun when driving a car on a highway. Fig.~\ref{fig:tunnel_frame}
    grayscale frame from a DAVIS camera. Fig.~\ref{fig:tunnel_megadepth} Depth
    prediction from MegaDepth. Fig.~\ref{fig:tunnel_events} Events and
    Fig.~\ref{fig:tunnel_prediction} our depth map prediction using events}
   \label{fig:tunnel_result}
\end{figure*}

\begin{figure*}[t!]
    \centering
    $\begin{array}{cccc}
    \subcaptionbox{\label{fig:e2vid} E2VID\cite{Rebecq19pami}}
    {
        \includegraphics[width=0.25\textwidth,keepaspectratio]{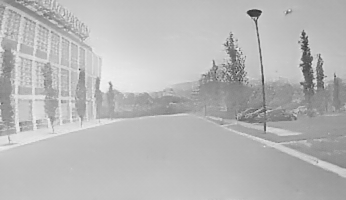}
    }
    \hspace{-0.12cm}
    \subcaptionbox{\label{fig:megadepth_e2vid} MegaDepth on E2VID}
    {
        \includegraphics[width=0.25\textwidth,keepaspectratio]{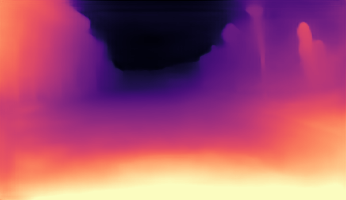}
    }
    \hspace{-0.12cm}
    \subcaptionbox{\label{fig:qualitative_events} Events}
    {
        \includegraphics[width=0.25\textwidth,keepaspectratio]{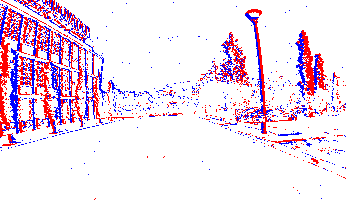}
    }
    \hspace{-0.12cm}
    \subcaptionbox{\label{fig:qualitative_prediction} Ours on events}
    {
        \includegraphics[width=0.25\textwidth,keepaspectratio]{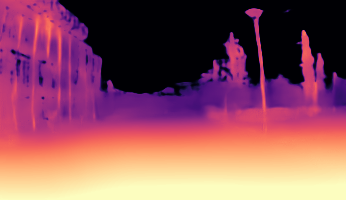}
    }
    \end{array}$
    \caption{Qualitative results comparing depth prediction using MegaDepth on reconstructed
    frames from E2VID and our method using events.}
   \label{fig:e2vid_comparison}
\end{figure*}

State of the art research in computer vision has demonstrated that salient edges
and texture are more relevant than color to estimate depth~\cite{huiccv19}.
Event cameras capture salient edges as well as detailed changes in brightness
with high temporal-spatial resolution. This makes event cameras suitable sensors
to predict depth.  We present two qualitative cases in this Appendix.
Fig.~\ref{fig:tunnel_result} shows a comparison by estimating depth from events
in an HDR scenario. Our method predicts depth directly from events while MegaDepth
uses the greyscale frame. MegaDepth shows difficulties to mask the sky due to
overexposure while the prediction from events better approximates the structure
of the scene. However, our method has some difficulties to predict the truck on
the right side of the image. This is because our DENSE dataset and
the~\textit{outdoor day2} sequence from MVSEC do not have moving trucks, so the
network has never seen such a situation during training.
Fig.~\ref{fig:e2vid_comparison} depicts the case of predicting depth from a
reconstructed frame using E2VID~\cite{Rebecq19pami}, which has HDR and does not
suffer from motion blur. The reconstruction shows that depth prediction directly
from events has more level of detail than from a reconstructed frame.

{\small
\balance
\bibliographystyle{ieee}
\bibliography{all}

\begin{thebibliography}{10}\itemsep=-1pt

\bibitem{Saxena06nips}
S.~H.~C. A.~Saxena and A.~Y. Ng.
\newblock Learning depth from single monocular images.

\bibitem{Acharta12pami}
R.~{Achanta}, A.~{Shaji}, K.~{Smith}, A.~{Lucchi}, P.~{Fua}, and
  S.~{Süsstrunk}.
\newblock Slic superpixels compared to state-of-the-art superpixel methods.
\newblock {\em IEEE Transactions on Pattern Analysis and Machine Intelligence},
  34(11):2274--2282, 2012.

\bibitem{Saxena09pami}
M.~S. Ashutosh~Saxena and A.~Y. Ng.
\newblock Make3d: Learning 3d scene structure from asingle still image.
\newblock 2009.

\bibitem{Dosovitskiy17corl}
A.~Dosovitskiy, G.~Ros, F.~Codevilla, A.~Lopez, and V.~Koltun.
\newblock {CARLA}: {An} open urban driving simulator.
\newblock In {\em Conf. on Robotics Learning (CoRL)}, 2017.

\bibitem{eigen2015predicting}
D.~Eigen and R.~Fergus.
\newblock Predicting depth, surface normals and semantic labels with a common
  multi-scale convolutional architecture.
\newblock In {\em Proceedings of the IEEE international conference on computer
  vision}, pages 2650--2658, 2015.

\bibitem{fu2018deep}
H.~Fu, M.~Gong, C.~Wang, K.~Batmanghelich, and D.~Tao.
\newblock Deep ordinal regression network for monocular depth estimation.
\newblock In {\em Proceedings of the IEEE Conference on Computer Vision and
  Pattern Recognition}, pages 2002--2011, 2018.

\bibitem{Gallego20pami}
G.~Gallego, T.~Delbruck, G.~Orchard, C.~Bartolozzi, B.~Taba, A.~Censi,
  S.~Leutenegger, A.~Davison, J.~Conradt, K.~Daniilidis, and D.~Scaramuzza.
\newblock Event-based vision: {A} survey.
\newblock {\em {IEEE} Trans. Pattern Anal. Mach. Intell.}, 2020.

\bibitem{Gehrig19iccv}
D.~Gehrig, A.~Loquercio, K.~G. Derpanis, and D.~Scaramuzza.
\newblock End-to-end learning of representations for asynchronous event-based
  data.
\newblock In {\em Int. Conf. Comput. Vis. (ICCV)}, 2019.

\bibitem{Geiger13ijrr}
A.~Geiger, P.~Lenz, C.~Stiller, and R.~Urtasun.
\newblock Vision meets robotics: The {KITTI} dataset.
\newblock {\em Int. J. Robot. Research}, 32(11):1231--1237, 2013.

\bibitem{Godard17cvpr}
C.~Godard, O.~{Mac Aodha}, and G.~J. Brostow.
\newblock Unsupervised monocular depth estimation with left-right consistency.
\newblock In {\em {IEEE} Conf. Comput. Vis. Pattern Recog. (CVPR)}, 2017.

\bibitem{godard2019digging}
C.~Godard, O.~Mac~Aodha, M.~Firman, and G.~J. Brostow.
\newblock Digging into self-supervised monocular depth estimation.
\newblock In {\em Proceedings of the IEEE International Conference on Computer
  Vision}, pages 3828--3838, 2019.

\bibitem{huiccv19}
J.~Hu, Y.~Zhang, and T.~Okatani.
\newblock Visualization of convolutional neural networks for monocular depth
  estimation, October 2019.

\bibitem{Ioffe15icml}
S.~Ioffe and C.~Szegedy.
\newblock Batch normalization: Accelerating deep network training by reducing
  internal covariate shift.
\newblock In {\em Proc. Int. Conf. Mach. Learning (ICML)}, 2015.

\bibitem{Kim14bmvc}
H.~Kim, A.~Handa, R.~Benosman, S.-H. Ieng, and A.~J. Davison.
\newblock Simultaneous mosaicing and tracking with an event camera.
\newblock In {\em British Mach. Vis. Conf. (BMVC)}, 2014.

\bibitem{Kim16eccv}
H.~Kim, S.~Leutenegger, and A.~J. Davison.
\newblock Real-time {3D} reconstruction and 6-{DoF} tracking with an event
  camera.
\newblock In {\em Eur. Conf. Comput. Vis. (ECCV)}, pages 349--364, 2016.

\bibitem{Kingma15iclr}
D.~P. Kingma and J.~L. Ba.
\newblock Adam: A method for stochastic optimization.
\newblock {\em Int. Conf. Learn. Representations ({ICLR})}, 2015.

\bibitem{Lenz15ijrr}
I.~Lenz, H.~Lee, and A.~Saxena.
\newblock Deep learning for detecting robotic grasps.
\newblock {\em Int. J. Robot. Research}, 34(4-5):705--724, 2015.

\bibitem{Li18cvpr}
Z.~Li and N.~Snavely.
\newblock Megadepth: Learning single-view depth prediction from internet
  photos.
\newblock In {\em {IEEE} Conf. Comput. Vis. Pattern Recog. (CVPR)}, 2018.

\bibitem{Lichtsteiner08ssc}
P.~Lichtsteiner, C.~Posch, and T.~Delbruck.
\newblock {A 128$\times$128 120 dB 15 $\mu$s latency asynchronous temporal
  contrast vision sensor}.
\newblock {\em {IEEE} J. Solid-State Circuits}, 43(2):566--576, 2008.

\bibitem{Silberman12eccv}
P.~K. Nathan~Silberman, Derek~Hoiem and R.~Fergus.
\newblock Indoor segmentation and support inference from rgbd images.
\newblock In {\em ECCV}, 2012.

\bibitem{Posch10isscc}
C.~Posch, D.~Matolin, and R.~Wohlgenannt.
\newblock A {QVGA} 143{dB} dynamic range asynchronous address-event {PWM}
  dynamic image sensor with lossless pixel-level video compression.
\newblock In {\em {IEEE} Intl. Solid-State Circuits Conf. (ISSCC)}, pages
  400--401, 2010.

\bibitem{Rebecq16bmvc}
H.~Rebecq, G.~Gallego, and D.~Scaramuzza.
\newblock {EMVS}: Event-based multi-view stereo.
\newblock In {\em British Mach. Vis. Conf. (BMVC)}, 2016.

\bibitem{Rebecq18corl}
H.~Rebecq, D.~Gehrig, and D.~Scaramuzza.
\newblock {ESIM}: an open event camera simulator.
\newblock In {\em Conf. on Robotics Learning (CoRL)}, 2018.

\bibitem{Rebecq17bmvc}
H.~Rebecq, T.~Horstschaefer, and D.~Scaramuzza.
\newblock Real-time visual-inertial odometry for event cameras using
  keyframe-based nonlinear optimization.
\newblock In {\em British Mach. Vis. Conf. (BMVC)}, 2017.

\bibitem{Rebecq17ral}
H.~Rebecq, T.~Horstsch{\"a}fer, G.~Gallego, and D.~Scaramuzza.
\newblock {EVO}: A geometric approach to event-based 6-{DOF} parallel tracking
  and mapping in real-time.
\newblock {\em {IEEE} Robot. Autom. Lett.}, 2(2):593--600, 2017.

\bibitem{Rebecq19cvpr}
H.~Rebecq, R.~Ranftl, V.~Koltun, and D.~Scaramuzza.
\newblock Events-to-video: Bringing modern computer vision to event cameras.
\newblock In {\em {IEEE} Conf. Comput. Vis. Pattern Recog. (CVPR)}, 2019.

\bibitem{Rebecq19pami}
H.~Rebecq, R.~Ranftl, V.~Koltun, and D.~Scaramuzza.
\newblock High speed and high dynamic range video with an event camera.
\newblock {\em {IEEE} Trans. Pattern Anal. Mach. Intell.}, 2019.

\bibitem{Ronneberger15icmicci}
O.~Ronneberger, P.~Fischer, and T.~Brox.
\newblock {U}-net: Convolutional networks for biomedical image segmentation.
\newblock In {\em International Conference on Medical Image Computing and
  Computer-Assisted Intervention}, 2015.

\bibitem{Rosinol18ral}
A.~{Rosinol Vidal}, H.~Rebecq, T.~Horstschaefer, and D.~Scaramuzza.
\newblock Ultimate {SLAM}? combining events, images, and {IMU} for robust
  visual {SLAM} in {HDR} and high speed scenarios.
\newblock {\em {IEEE} Robot. Autom. Lett.}, 3(2):994--1001, Apr. 2018.

\bibitem{Scheerlinck18accv}
C.~Scheerlinck, N.~Barnes, and R.~Mahony.
\newblock Continuous-time intensity estimation using event cameras.
\newblock In {\em Asian Conf. Comput. Vis. (ACCV)}, 2018.

\bibitem{Scheerlinck20wacv}
C.~Scheerlinck, H.~Rebecq, D.~Gehrig, N.~Barnes, R.~Mahony, and D.~Scaramuzza.
\newblock Fast image reconstruction with an event camera.
\newblock In {\em {IEEE} Winter Conf. Appl. Comput. Vis. (WACV)}, 2020.

\bibitem{Shi15nips}
X.~Shi, Z.~Chen, H.~Wang, D.~Yeung, W.~Wong, and W.~Woo.
\newblock Convolutional {LSTM} network: {A} machine learning approach for
  precipitation nowcasting.
\newblock In {\em Conf. Neural Inf. Process. Syst. (NIPS)}, 2015.

\bibitem{Tulyakov19iccv}
S.~Tulyakov, F.~Fleuret, M.~Kiefel, P.~Gehler, and M.~Hirsch.
\newblock Learning an event sequence embedding for dense event-based deep
  stereo.
\newblock In {\em Int. Conf. Comput. Vis. (ICCV)}, 2019.

\bibitem{wang2018learning}
C.~Wang, J.~Miguel~Buenaposada, R.~Zhu, and S.~Lucey.
\newblock Learning depth from monocular videos using direct methods.
\newblock In {\em Proceedings of the IEEE Conference on Computer Vision and
  Pattern Recognition}, pages 2022--2030, 2018.

\bibitem{zhou2017unsupervised}
T.~Zhou, M.~Brown, N.~Snavely, and D.~G. Lowe.
\newblock Unsupervised learning of depth and ego-motion from video.
\newblock In {\em Proceedings of the IEEE Conference on Computer Vision and
  Pattern Recognition}, pages 1851--1858, 2017.

\bibitem{Zhou18eccv}
Y.~Zhou, G.~Gallego, H.~Rebecq, L.~Kneip, H.~Li, and D.~Scaramuzza.
\newblock Semi-dense {3D} reconstruction with a stereo event camera.
\newblock In {\em Eur. Conf. Comput. Vis. (ECCV)}, pages 242--258, 2018.

\bibitem{Zhu17cvpr}
A.~Z. Zhu, N.~Atanasov, and K.~Daniilidis.
\newblock Event-based visual inertial odometry.
\newblock In {\em {IEEE} Conf. Comput. Vis. Pattern Recog. (CVPR)}, pages
  5816--5824, 2017.

\bibitem{Zhu18eccv}
A.~Z. Zhu, Y.~Chen, and K.~Daniilidis.
\newblock Realtime time synchronized event-based stereo.
\newblock In {\em Eur. Conf. Comput. Vis. (ECCV)}, pages 438--452, 2018.

\bibitem{Zhu18ral}
A.~Z. Zhu, D.~Thakur, T.~Ozaslan, B.~Pfrommer, V.~Kumar, and K.~Daniilidis.
\newblock The multivehicle stereo event camera dataset: An event camera dataset
  for {3D} perception.
\newblock {\em {IEEE} Robot. Autom. Lett.}, 3(3):2032--2039, July 2018.

\bibitem{Zhu19cvpr}
A.~Z. Zhu, L.~Yuan, K.~Chaney, and K.~Daniilidis.
\newblock Unsupervised event-based learning of optical flow, depth, and
  egomotion.
\newblock In {\em {IEEE} Conf. Comput. Vis. Pattern Recog. (CVPR)}, 2019.

\end{thebibliography}
}

\end{document}